\documentclass[sigconf]{acmart}

\AtBeginDocument{%
  }

\setcopyright{acmlicensed}
\copyrightyear{2024}
\acmYear{2024}
\acmDOI{10.1145/3627673.3679795}

\acmConference[CIKM '24]{Proceedings of the 33rd ACM International Conference on Information and Knowledge Management}{October 21--25, 2024}{Boise, ID, USA}
\acmBooktitle{Proceedings of the 33rd ACM International Conference on Information and Knowledge Management (CIKM '24), October 21--25, 2024, Boise, ID, USA}

\acmISBN{979-8-4007-0436-9/24/10}

\usepackage{algorithm}
\usepackage{algpseudocode}
\usepackage[normalem]{ulem}
\usepackage{graphicx}
\usepackage{subcaption}
\usepackage{balance}

\usepackage{array}
\usepackage{booktabs}
\usepackage{amsmath}

\newtheorem{definition}{Definition}

\DeclareMathOperator{\similarity}{sim}

\begin{document}

\title{Semantic Prototypes: Enhancing Transparency without Black Boxes}

\author{Orfeas Menis-Mastromichalakis}
\email{menorf@ails.ece.ntua.gr}
\affiliation{%
  \department{Artificial Intelligence and Learning Systems Laboratory}
  \institution{National Technical University of Athens}
   \city{Athens}
   \country{Greece}
}

\author{Giorgos Filandrianos}
\email{geofila@islab.ntua.gr}
\affiliation{%
  \department{Artificial Intelligence and Learning Systems Laboratory}
  \institution{National Technical University of Athens}
   \city{Athens}
   \country{Greece}
}

\author{Jason Liartis}
\email{jliartis@ails.ece.ntua.gr}
\affiliation{%
  \department{Artificial Intelligence and Learning Systems Laboratory}
  \institution{National Technical University of Athens}
   \city{Athens}
   \country{Greece}
}

\author{Edmund Dervakos}
\email{eddiedervakos@islab.ntua.gr}
\affiliation{%
    \department{Artificial Intelligence and Learning Systems Laboratory}
  \institution{National Technical University of Athens}
   \city{Athens}
   \country{Greece}
}

\author{Giorgos Stamou}
\email{gstam@cs.ntua.gr}
\affiliation{%
  \department{Artificial Intelligence and Learning Systems Laboratory}
  \institution{National Technical University of Athens}
   \city{Athens}
   \country{Greece}
}


\begin{abstract}
  As machine learning (ML) models and datasets increase in complexity, the demand for methods that enhance explainability and interpretability becomes paramount. Prototypes, by encapsulating essential characteristics within data, offer insights that enable tactical decision-making and enhance transparency. Traditional prototype methods often rely on sub-symbolic raw data and opaque latent spaces, reducing explainability and increasing the risk of misinterpretations. This paper presents a novel framework that utilizes semantic descriptions to define prototypes and provide clear explanations, effectively addressing the shortcomings of conventional methods. Our approach leverages concept-based descriptions to cluster data on the semantic level, ensuring that prototypes not only represent underlying properties intuitively but are also straightforward to interpret. Our method simplifies the interpretative process and effectively bridges the gap between complex data structures and human cognitive processes, thereby enhancing transparency and fostering trust. Our approach outperforms existing widely-used prototype methods in facilitating human understanding and informativeness, as validated through a user survey. 
\end{abstract}

\ccsdesc[500]{Information systems~Information retrieval}
\ccsdesc[500]{Computing methodologies~Knowledge representation and reasoning}
\keywords{Explainability, Transparency, Semantic, Prototypes, Interpretability}
\begin{teaserfigure}
  \includegraphics[width=\textwidth]{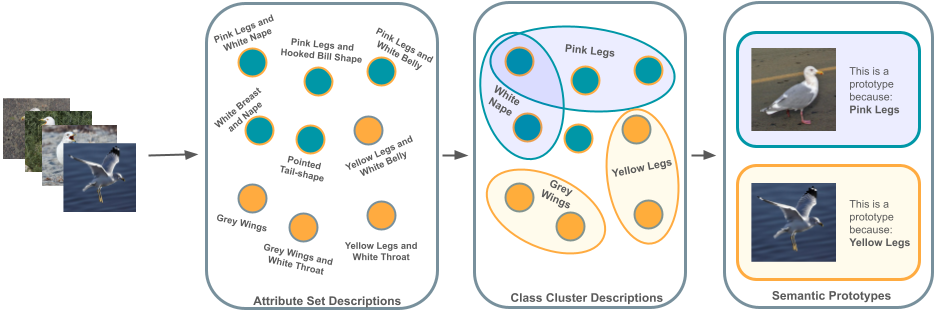}
  \caption{Overview of our Method.}
  \Description{Sample Description}
  \label{fig:teaser}
\end{teaserfigure}


\maketitle

\section{Introduction}
\label{sec:intro}

In the rapidly evolving landscape of data-driven decision-making and machine learning (ML) advancements, the pursuit of explainability and interpretability stands as a critical imperative. As ML models evolve in complexity and scope, understanding their decision-making processes becomes paramount for fostering trust, ensuring accountability, and promoting fairness. Equally crucial is the comprehension of the datasets upon which these models are trained and applied. Data, often vast and heterogeneous, serve as the foundational bedrock upon which ML models operate. However, the sheer volume and intricacy of data present formidable challenges in discerning meaningful insights, uncovering hidden biases, and ensuring the quality and fairness of AI-driven systems. Thus, the need for transparent and interpretable methodologies that not only shed light on ML model behavior but also facilitate a deeper understanding and management of data is unequivocal.

Prototypes have emerged as pivotal constructs not only for explaining machine learning models but also for comprehending the underlying data \cite{kim2016examples}. Acting as archetypal representations, prototypes encapsulate the essential characteristics or features of specific clusters or classes within a dataset, providing intuitive insights into its inherent properties. Research on human cognition and reasoning has shown that the use of prototypical examples is fundamental to the development of effective strategies for tactical decision-making \cite{newell1972human, cohen1996metarecognition}, and recent user studies show that concept-based and prototype explanations are prefered by users over other existing explanations \cite{kim2023help}.
For instance, in information retrieval, prototypes act as exemplars for enhancing search efficiency and relevance ranking by aiding in query expansion. Additionally, there is a growing interest within the AI community in case-based reasoning and prototype-based classifiers, highlighting the versatility and acceptance of prototypes in various applications. By leveraging prototypes, stakeholders can navigate the complexities of data-driven decision-making more effectively, fostering transparency and enabling nuanced decision-making processes.

\begin{figure}[t]
  \centering
  \includegraphics[width=\linewidth]{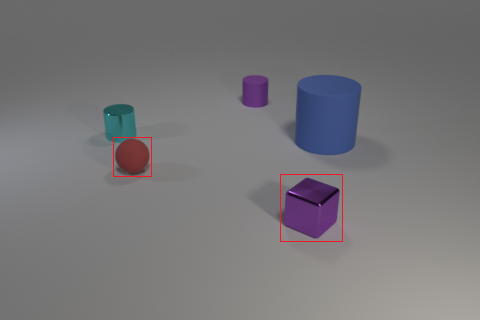}
  \caption{A sample image from class 2 of the CLEVR-Hans3 dataset.}
  \label{fig:clevr-intro}
  \Description{A sample image from class 2 of the CLEVR-Hans3 dataset, depicting a big blue rubber cylinder, a small purple rubber cylinder, a small cyan metal cylinder, a small red rubber sphere, and a small purple metal cube.}
\end{figure}

However, the majority of existing prototype approaches exhibit a major structural limitation that undermines their effectiveness and trustworthiness: they rely solely on the raw, unstructured feature space. This can be problematic from many aspects. Firstly, the feature space is often not understandable, an issue that persists across many eXplainable AI (XAI) methods, and can result in lack of intuition and potential for misinformation \cite{rudin2019stop, mittelstadt2019explaining, mastromichalakis2024rule, liartis2023searching, miller2019explanation}. For example consider genomics, where the feature space consists of DNA sequences which can consist of millions or billions of base pairs for an organism, and in which there can be interdependencies that are thousands of base pairs apart. Parts of the genome might be irrelevant, others might regulate the expression of genes that are elsewhere in the sequence, others might be genes themselves etc. This raw feature representation is not understandable, even to a domain expert, so a traditional prototype in this case would not be helpful.
Secondly, in many models, especially those involving complex interactions or relationships between features, a single or even a few examples might not capture the full range of interactions or the subtleties involved in model decisions. This can make it difficult to convey the full complexity of the model's decision-making process through just prototypical examples, and can lead to oversimplification and misinterpretation. For example, consider the image shown in Figure \ref{fig:clevr-intro} from the CLEVR-Hans3 dataset \cite{stammer2021right}, for which it is known that the class semantics are ``small metal cube and small sphere''. By just looking at the pixel representation of the prototype, even if the prototypical parts of the image have been highlighted (see bounding boxes in Figure \ref{fig:clevr-intro}), it is impossible to discern the characteristics that make them prototypical. It could be the color, size, shape or texture of each object, or even their location in the image. Therefore, without telling a user the specific semantics that make this image (and the highlighted parts) prototypical, it is easy for them to misinterpret the explanation and be misled. 
Thirdly, prototypes cannot be expected to generalize to all cases, and even though they might be excellent representations of a particular class, it is not made clear to an end-user which aspects of the prototype make it representative, and to which cases it might generalize.
Additionally, several prototype methods do not act on the feature representation itself, opting instead to utilize black-box models that transform the features into a lower-dimensional latent space representation
. This exacerbates the aforementioned issues, as latent representations are non-intuitive and unintelligible to humans  \cite{wan2024interpretable}, and it also creates a paradoxical situation where non-interpretable models are used to provide explanations or interpretability, which might also facilitate malicious manipulations \cite{hoffmann2021looks}. Instead, recent research emphasizes the importance of \emph{explaining the prototypes} \cite{nauta2021looks, wan2024interpretable} underscoring the necessity for a semantic level of information alongside prototypes.

Our approach represents a novel solution to address the limitations in existing prototype methods. To tackle the challenge of using raw data to define prototypes, we propose a shift towards semantic prototypes. In our approach, prototypes are not selected based on raw input features but on the semantic descriptions associated with each data point. By leveraging concept-based semantic descriptions to create clusters of data described by semantic rules as shown in Figure~\ref{fig:teaser}, we ensure that prototypes are representative of the underlying data distribution while maintaining transparency and interpretability. This process eliminates the need to map data to a non-interpretable latent space, as distances are measured on the more intuitive semantic level. Moreover, our method dynamically determines the number of prototypes needed to semantically cover the entire data distribution, enhancing its adaptability and effectiveness. Furthermore, our approach mitigates the issue of providing explanations solely in terms of raw sub-symbolic data by providing both prototypical examples and the corresponding prototypical semantic description. Each cluster's semantic description serves as the prototypical part of the data point on the semantic level, offering insights into why a particular example is deemed a prototype. This combination of prototypical examples and semantic descriptions bridges the gap between the semantic and data levels, enhancing the interpretability and trustworthiness of our method. By enabling users to question and scrutinize each step of the process on the semantic level, we foster warranted trust and confidence in the explanations provided. Thus, our approach offers a simple yet effective and intuitive solution to enhance the interpretability of prototype-based explanations, mitigating the drawbacks of existing approaches.

\section{Related Work}
Our work is positioned at the intersection of several key areas within artificial intelligence, notably explainable AI (XAI), prototype-based methods, and case-based reasoning \cite{aamodt1994case}. By leveraging semantic prototypes, our approach not only enhances model interpretability but also facilitates a clearer understanding of datasets, offering a comprehensive overview of their inherent structure and characteristics. 
Classical algorithms like k-medoids \cite{rdusseeun1987clustering} have traditionally been used to select representative subsets of data points, illustrating early methods of data summarization through clustering. More recently, seminal works such as \cite{kim2016examples} and \cite{gurumoorthy2019efficient} have leveraged prototypes to critically evaluate models and enhance transparency in machine learning decisions, establishing prototypes as an interpretability method.

A significant body of research has focused on using prototypes to create more interpretable classifiers. This approach, often referred to as case-based or example-based classification, aims to enhance the transparency of AI models by relying on representative examples. By integrating prototypes into the classification process, these methods strive to provide intuitive, example-driven explanations that make the model's decisions more understandable to humans. \cite{chen2019looks} is a seminal work in this direction, introducing ProtoPNet, a deep network architecture that dissects images by finding prototypical parts and combines evidence from these prototypes to make final classifications. \cite{wang2023learning} claims to improve the classification performance of ProtoPNet with a method to learn support prototypes that lie near the classification boundary in the feature space. In a similar vein, \cite{nauta2021neural} introduces ProtoTree, a prototype-based decision tree that employs prototypical patches as proxies for class-defining semantics. 
Several other works follow this rationale of prototypical learning through various approaches \cite{angelov2020towards, arik2020protoattend, xue2022protopformer, li2018deep, rymarczyk2021protopshare, rymarczyk2022interpretable, wang2023learning}. However, their reliance on raw data limits the interpretability of their methods and the intuitiveness of the prototypes, potentially leading to misleading explanations.

Recent research aligns with our work by acknowledging the limitations of providing explanations in terms of raw data and highlights the necessity to ``explain the prototypes''. In \cite{nauta2021looks}, the authors introduce a method to provide further insights for prototypes based on existing methods like ProtoPNet by altering some characteristics of the image, such as hue and saturation, and providing explanations based on that information. Similarly, \cite{wan2024interpretable} proposes the Semantic Prototype Analysis Network (SPANet), an interpretable object recognition approach that through additional semantic information enables models to explicate the decision process by ``pointing out where to focus'' and ``explaining why'' on a semantic level. Our work also utilizes information on the semantic level to define prototypes and produce semantically enriched explanations, bearing strong similarities with recent rule-based \cite{mastromichalakis2024rule, liartis2023searching, liartis2021semantic} and counterfactual methods \cite{dervakos2023choose} that use semantic descriptions of data to provide intuitive, human-understandable explanations.

Our work is further supported by research discussing the shortcomings of latent space prototype interpretability, as outlined in \cite{hoffmann2021looks}, where the non-interpretability of latent space is highlighted, showing that existing methods using non-interpretable embedding spaces limit the interpretability of prototypes and are vulnerable to malicious attacks. Efforts to bridge the ``semantic gap'' between latent space and pixel space through the correlation of prototypes with ground-truth object parts \cite{nauta2023pip} still rely on opaque procedures to map raw data to the latent space. \cite{wang2021interpretable} claims to address this opacity by introducing a plug-in transparent embedding space to bridge high-level input patches and output categories.

In contrast, our approach eliminates the reliance on non interpretable latent spaces by using semantic descriptions directly, making each step transparent and interpretable. This not only enhances trust in the explanations provided but also allows for a more robust understanding of both the model and the data. Our method stands out by addressing both the need for clear, semantic-level explanations and the requirement for prototypes that truly represent the data in an intuitive and human-understandable way.

\section{Semantic Prototypes}

In this section we define the proposed framework for semantic prototypes. At the core of the framework is the notion of an \emph{Attribute Set Description (ASD)}, which provides a simple way to represent data samples semantically, as a set of entities, where an entity is represented as a set of attributes. 

\begin{definition}[Attribute Set Description]
Given a set $\mathcal{V}$, an Attribute Set Description is a set of the form $\{s_1, s_2, \dots, s_n\}$ where each $s_i$ is of the form $\{a_{i1}, a_{i2}, \dots, a_{im_i}\} \subseteq \mathcal{V}$.
\end{definition}

The set $\mathcal{V}$ is a vocabulary that lists all the possible attributes an entity can have, so an {ASD} lists the attributes of a collection of entities. For defining semantic prototypes, we assume that we have data where samples are described by an {ASD}.
Specifically, we assume that our data $D$ consist of triples $d = (x, y, z)$ where $x$ is a raw data point, (e.g. an image, audio signal, DNA sequence etc.) $y$ is a label, and $z$ an semantic description of that data point. We assume that $z$ is an ASD that reflects the contents of $x$.
In the case of an image the entities could be objects depicted within the image, each characterised by shape, colour, size, etc., while in the case of a speech signal, the entities could be utterances that are characterized by loudness, pitch, intonation, rhythm etc.

The assumptions that i) there are available data with {ASDs} and ii) that the {ASDs} accurately describe the data samples are worth further discussion. In the ideal case, such semantic descriptions will have resulted by human expert annotation, especially in decision-critical domains. {There already exist multiple datasets with manually-added semantic descriptions or metadata that can be used as ASDs, both for general-purpose and domain-specific tasks, such as Audio set \cite{gemmeke2017audio} including audio events accompanied by an ontology describing their semantics, the Visual Genome \cite{krishna2017visual}, containing images accompanied by scene graphs, where entities are linked semantically to WordNet \cite{fellbaum2010wordnet}, and the cancer genome atlas  \cite{weinstein2013cancer}, that includes genomic sequences along with a rich set of clinical and other information, among others.} Furthermore, one could also use traditional, transparent feature extraction techniques to generate the {ASDs}, or, even more complex models, such as large vision-language models, {similar to recent works that relate to ours \cite{wan2024interpretable}. The point of the {ASD} is to provide a meaningful description of a data sample at a level of abstraction that is understandable. 


An {ASD} can also be used to describe a set of data samples. Given an ASD $z'$, we will say that $z'$ subsumes $z$ if $\forall s_i \in z' ~~\exists s_j \in z ~~s.t. ~s_i \subseteq s_j$. This can be thought of as $z'$ being more general than $z$. Given a data point $d = (x, y, z)$, if $z'$ subsumes $z$ we will also say that $z'$ describes the data point $d$. Essentially, $z'$ describes $d$, if the description $z$ of $d$ contains entities with attributes that match or exceed those described in $z'$, thus, there can be {ASDs} that describe multiple data points.  {For example a data sample with ASD $\{\{\mathsf{Cat}\},\{\mathsf{Dog}\}\}$ is described by the ASD $\{\{\mathsf{Cat}\}\}$ , and so is the data sample with ASD $\{\{\mathsf{Cat}\},\{\mathsf{Mouse}\}\}$}. We utilize this idea for the semantic prototypes, by first finding {ASDs} which describe only data points with a particular label. We call such ASDs \emph{class {cluster} descriptions (CCD)}  of that label. 

\begin{definition}[Class  {Cluster} Description]
    A class {cluster} description of class $c$, is an ASD $r$ such that, if $r$ describes a datapoint $d = (x, y, z)$ (i.e. $r$ subsumes $z$), then $y = c$.
\end{definition}

{Intuitively, a CCD semantically describes a cluster of data points that belong to a specific class, and no other data points. It can be interpreted as an IF THEN rule in that IF a data point is described by a CCD, then it belongs to that particular class. The purpose of identifying and semantically describing clusters of data points is to subsequently find the most representative or informative samples for those clusters, which can then be given as prototypes, along with their semantic description (ASD), and the semantic description of why they belong to their class (CCD). In particular,} given a {CCD} for a label, the corresponding semantic prototype is the data point whose ASD contains the least redundant information among points that fit that description.





\begin{definition}[Semantic Prototype]
A semantic prototype $p$ for a class {cluster} description $r$ is a data point $p=(x,y,z)$ that is described by $r$, and for which, given a distance metric $\mathsf{dist}$, for every other data point $d'$ that is described by $r$, it holds that $\mathsf{dist}(z,r)\leq\mathsf{dist}{(z',r)}$. 
\end{definition}

Intuitively, this means that a semantic prototype is a data point that materialises all the semantic information of the class description, since it is described by it, and contains as little extra information as possible.
The choice to limit the extra information is made to ensure that end-users are not distracted by irrelevant characteristics of the data point, such as objects in an image that do not affect what class it belongs to. Regarding the distance metric $\mathsf{dist}$, in our implementation we opt for a \textit{set edit distance}, as it has been used in other semantic explainability methods \cite{dervakos2023choose}, but other distance metrics could potentially be used, such as the extension of Jaccard similarity to sets of sets.

\section{Computing Semantic Prototypes}


Within the proposed framework we can find prototypes using semantic criteria, and we can also answer the question "Why is this example a prototype?", by accompanying the prototypical example with a semantic class description when showing it to an end-user. 
To this end, there are are two main components that need to be computed. First, is the process of identifying and describing  clusters within each class (computing CCDs), and second is the process of choosing the most informative data sample for each cluster.


\subsection{Finding class cluster descriptions}

\begin{algorithm}
\caption{}
\label{alg:main}
\begin{algorithmic}[1]
\Require A set $\mathcal{P}$ of positive data points and a set $\mathcal{N}$ of negative data points.
\Ensure A set $C$ of class cluster descriptions.
\State $C \gets \emptyset$
\For {$d = (x, y, z) \in \mathcal{P}$}
    \State description $\gets z$
    \State $l \gets \mathcal{P} \setminus \{d\}$ as a list sorted by similarity to $z$
    \For {$d' = (x', y, z') \in l$}
        \State remove $d'$ from $l$
        \State ncd $\gets$ merge(description, $z'$) \Comment{new candidate description}
        \If {ncd does not describe any data point in $\mathcal{N}$}
            \State description $\gets$ ncd
        \EndIf
        \State $l \gets l$ sorted by similarity to description
    \EndFor
    \State $C \gets C \cup \{$description$\}$
\EndFor
\end{algorithmic}
\end{algorithm}

As the space of all possible CCDs is exponentially large, our approach works by first heuristically generating a large (but polynomial) number of potential CCDs, filtering out those that do not satisfy the criteria (i.e. the clusters contain data samples that have a different label), and finally choosing a subset of the computed CCDs, depending on the number of prototypes that we want to produce and on the class coverage of the CCDs.

Given a dataset $D$, and a class $c$ for which we want to produce prototypical examples, we would ideally like to produce the smallest number of {CCDs} that describe the entirety of class $c$ without describing any data points from other classes. {It is worth mentioning that since finding CCDs is equivalent to finding \emph{rules} of the form ``IF data sample contains entities with specific attributes THEN it is classified to class $c$'', existing rule-based methods could be adapted for finding CCDs \cite{zhou2003extracting,augasta2012reverse,mastromichalakis2024rule,liartis2021semantic}.}

In our implementation, we utilise Algorithm \ref{alg:main} to compute the initial {CCDs}, using $\mathcal{P} = \{(x,y,z) \in D: y = c\}$ as the positive data points and $\mathcal{N} = \{(x,y,z) \in D: y \neq c\}$ as the set of negative data points. Alg. \ref{alg:main} is a greedy algorithm inspired by \cite{liartis2021semantic} that starts with an ASD, and using a similarity metric (eq. \ref{eq:sim}) as a heuristic, greedily merges (Alg. \ref{alg:merge}) positive descriptions into more general ones, and subsequently checks that the more general descriptions do not describe any negative data points. In contrast to \cite{liartis2021semantic}, we repeat this process for each positive data point, because we want to ensure that each positive data point fits at least one {CCD}, and to also mitigate "bad" choices induced by the heuristic. This strikes a balance between only utilizing each data point once, and exploring the combinatorially large number of all possible choices.

\begin{equation}
\begin{split}
    \similarity(z_1, z_2) =~ &\frac{1}{2}\sum_{s_i \in z_1} \max_{s_j \in z_2} |s_i \cap s_j|/(|z_1|\cdot|s_i \cup s_j|) \\
    +~ &\frac{1}{2}\sum_{s_i \in z_2} \max_{s_j \in z_1} |s_i \cap s_j|/(|z_2|\cdot|s_i \cup s_j|)
\end{split}
\label{eq:sim}
\end{equation}


The similarity metric we use, as described in equation \ref{eq:sim}, utilizes the Jaccard similarity to compare the entities described in each ASD. For each entity in $z_1$, it calculates the maximum number of attributes it shares with any entity in $z_2$. This is averaged over the entities in $z_1$, repeated symmetrically for the entities in $z_2$, and then these two quantities are averaged. We average over entities so that if $z_1$ describes many more entities than $z_2$, it does not dominate the total similarity and vice versa. The two quantities are averaged so that the final quantity is between 0 and 1, as is commonly required of a similarity metric.

\begin{algorithm}
\caption{merge}
\label{alg:merge}
\begin{algorithmic}
\Require Two ASDs $z_1$, $z_2$.
\Ensure An ASD $z$ which generalizes $z_1$ and $z_2$.
\State $z' \gets \{s_i \cap s_j~|~s_i \in z_1, s_j \in z_2\}$ \Comment combine
\State $z \gets \{s_i \in z'~|~\nexists s_j \in z' \textnormal{ s.t. } s_i \subset s_j\}$ \Comment trim
\end{algorithmic}
\end{algorithm}

The merge operation also follows the paradigm of \cite{liartis2023searching}, by finding all common attributes for pairs of entities from $z_1$ and $z_2$, and then trims the resulting ASD. This way of combining $z_1$ and $z_2$ is essentially the direct product of finite structures, applied to ASDs. It is also the join operation on the lattice induced by ASD subsumption. The resulting ASD of this merge operation holds the property that it subsumes $z_1$ and $z_2$, and is subsumed by any other ASD that subsumes both $z_1$ and $z_2$. Therefore, it is their most specific generalization. This operation is widely adopted for separating structured positive and negative examples \cite{liartis2023searching, cima2022separability, GCAI2019:Ontology_Mediated_Queries_from_Examples, ijcai2018p255}. The trimming operation used only removes redundant entity descriptions, without sacrificing the property of the most specific generalization.

Alg.1 results in a large set of {CCDs} which are used as candidates for subsequently finding semantic prototypes. As each {CCD} results in a prototype, we want to limit their number so that they can all be shown to a user without overwhelming them. In our implementation we again do this greedily, by choosing at each step the {CCD} that describes the most positive samples that are not already described by any previously selected class descriptions. Selecting the fewest number of {CCDs} that in total describe all positive data points is an instance of the set cover problem, while selecting $k$ {CCDs} that describe as many positive data points as possible is an instance of the maximum coverage problem. Both problems are NP-complete and the greedy algorithm is the best polynomial-time approximation, up to lower-order terms, unless P=NP \cite{vazirani2001approximation, 10.1145/285055.285059, dinur2014analytical}.

\subsection{Finding semantic prototypes}

Having established the {CCDs}, we proceed to find prototypes for each class by identifying the closest data point to each {CCD} that is simultaneously described by it. In our implementation, the closeness is determined through the set edit distance, a metric quantifying the distance between the CCD and the ASD of each data point. In particular, as we know that the {CCD} $r$ describes the data sample $d = (x,y,z)$, the only necessary edits to transform $r$ into $z$ are insertions of attributes into the sets contained in $r$. 
To do this, for every pair of sets $(s_r,s_z)$ where $s_r\in{r}$, $s_z\in{z}$ and $s_r\subseteq{s_z}$ we compute the number of insertions $|s_z\setminus{s_r}|$. Then the pairs $(s_r,s_z)$ are organized into a bipartite graph, where the weights of the edges are set to be the number of insertions computed previously. It is guaranteed that every $s_r$ will have at least one edge, since we know that $r$ describes $z$, meaning that $\forall s_r \in r ~~\exists s_z \in z ~~s.t. ~s_r \subseteq s_z$. 
{Finally, we compute the minimum number of additions required to transform $s_r$ to $s_z$, yielding the edit distance between the class description and the data point, by adapting the minimum weight full match algorithm, as used in \cite{filandrianos2022conceptual}.} This is computed for all data points, and then the semantic prototype is chosen to be the one with the least edit distance.



\section{Experiments}
In our experiments, we utilized two datasets to evaluate the effectiveness of our approach: the CLEVR-Hans dataset \cite{stammer2021right} and the CUB-200 dataset \cite{wah2011caltech}. The CLEVR-Hans dataset comprises artificial images featuring a varying number of objects with different sizes, shapes, colors, and textures. This dataset is chosen because of its simplicity and clear semantics and characteristics that allow for a straightforward demonstration of how our method excels where other explanation techniques fall short. The second dataset we employed is the CUB-200 dataset, which consists of real images of birds divided into 200 classes according to species. This dataset allows us to evaluate our method in real-life images. It is widely used in prototype-based methods, making it ideal for comparison. Our code is available here: \url{https://github.com/ails-lab/Semantic-Prototypes}. 

\subsection{Qualitative Evaluation}
\subsubsection{CLEVR-Hans}
We conducted experiments on the CLEVR-Hans dataset to qualitatively analyze the informativeness and interpretability of our semantic prototype approach. We compared our method against existing prototype-based techniques, focusing on how well each method captures the clear and distinct semantics present in the dataset. Each class in the dataset has a clear semantic description that characterizes all the images within that class. For example, all images in Class 1 contain at least one large cube and one large cylinder; Class 2 images feature at least one small metal cube and one small sphere; Class 3 images include at least one large blue sphere and one small yellow sphere. 
The following Class Characteristic Descriptions (CCDs) produced by our method correctly reflect the characteristics of each class  
$$
\mathsf{Class\_1}=\{\{\mathsf{Large},\mathsf{Cube}\}, \{\mathsf{Large},\mathsf{Cylinder}\}\}\\
$$
$$
\mathsf{Class\_2}=\{\{\mathsf{Small},\mathsf{Metal},\mathsf{Cube}\}, \{\mathsf{Small},\mathsf{Sphere}\}\}\\
$$
$$
\mathsf{Class\_3}=\{\{\mathsf{Large},\mathsf{Blue},\mathsf{Sphere}\}\{\mathsf{Small},\mathsf{Yellow},\mathsf{Sphere}\}\}\\
$$
Figure \ref{fig:six_images} shows the prototypical images for each class in the training set of CLEVR-Hans3, generated by Protodash \cite{gurumoorthy2019efficient} and our proposed approach. Our method selects prototypes with the least extraneous information, producing clearer, more focused prototypes that prevent cognitive overload and help users detect the distinguishing features between classes. By observing the prototypes produced by our method, users can more easily identify patterns due to the absence of distracting information. In contrast, prototypes generated by other methods often represent a ``central'' data point that may include irrelevant information.
Our approach essentially disregards the distribution of images in the feature space, where images with many objects might be more common. Instead, we find the best CCDs that cover the class as comprehensively as possible and then identify the data point described by the CCD with the fewest redundancies.

Additionally, our study highlights the importance of providing explanations alongside prototypes. Although our method minimizes irrelevant information in prototypes, extracting the actual semantics of the class remains challenging. This challenge is even more pronounced with prototypes produced by methods like Protodash, where the amount of encapsulated information can be overwhelming. Even when methods can detect the exact parts of images containing characteristic features, it can still be difficult to extract the correct semantics due to feature entanglement at the data level. As shown in Figure~\ref{fig:clevr-intro} and discussed in Section~\ref{sec:intro}, simply indicating the prototypical patch sometimes fails to clarify the prototypical characteristics due to this entanglement. Our method, through the use of CCDs, clearly presents the semantics of each class in a simple, intuitive, and informative manner.

\begin{figure}[t]
    \centering
    \begin{subfigure}[b]{0.45\linewidth}
        \includegraphics[width=\linewidth]{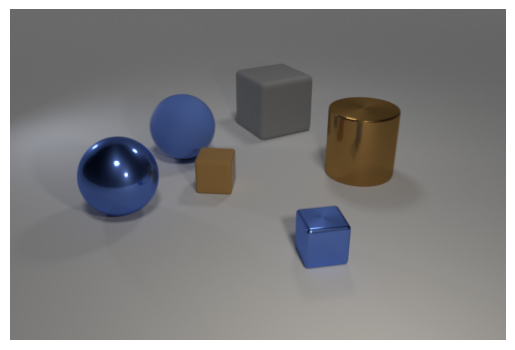}
        \caption{Protodash Class 1}
        \label{fig:image1}
    \end{subfigure}
    \hfill
    \begin{subfigure}[b]{0.45\linewidth}
        \includegraphics[width=\linewidth]{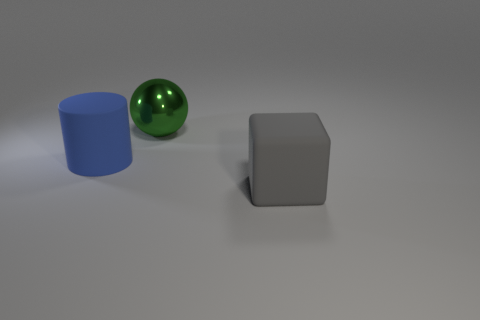}
        \caption{Ours Class 1}
    \end{subfigure}

    \vskip\baselineskip

    \begin{subfigure}[b]{0.45\linewidth}
        \includegraphics[width=\linewidth]{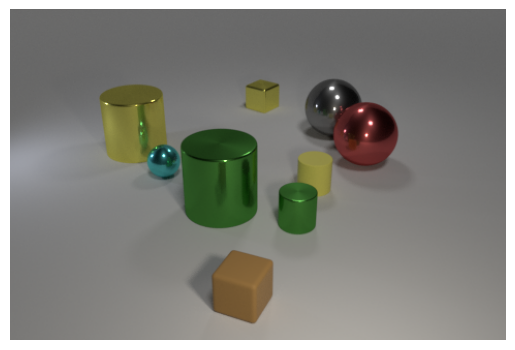}
        \caption{Protodash Class 2}
        \label{fig:image3}
    \end{subfigure}
    \hfill
    \begin{subfigure}[b]{0.45\linewidth}
        \includegraphics[width=\linewidth]{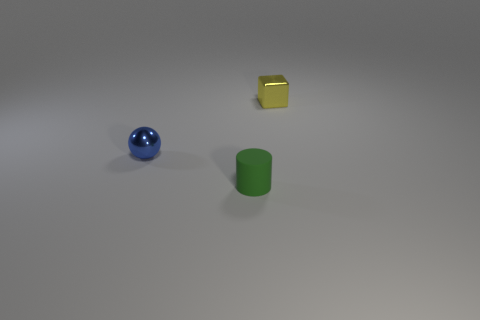}
        \caption{Ours Class 2}
        \label{fig:image4}
    \end{subfigure}

    \vskip\baselineskip

    \begin{subfigure}[b]{0.45\linewidth}
        \includegraphics[width=\linewidth]{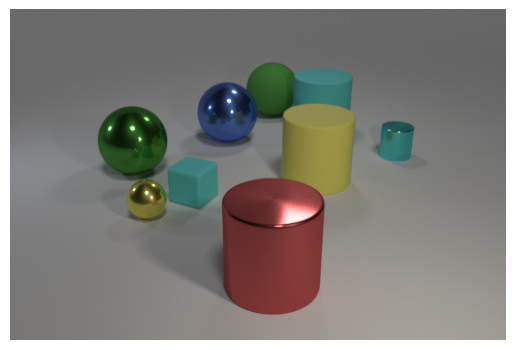}
        \caption{Protodash Class 3}
        \label{fig:image5}
    \end{subfigure}
    \hfill
    \begin{subfigure}[b]{0.45\linewidth}
        \includegraphics[width=\linewidth]{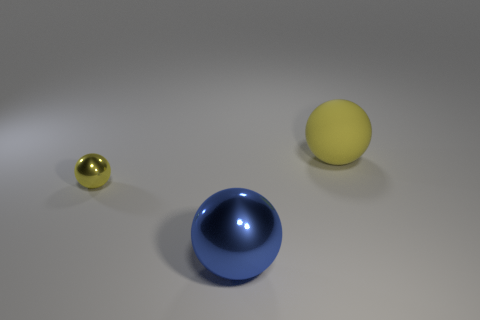}
        \caption{Ours Class 3}
        \label{fig:image6}
    \end{subfigure}

    \caption{CLEVR-Hans3 Prototypes}
    \label{fig:six_images}
\end{figure}

\subsubsection{CUB-200}
By analyzing the performance of our method on the CUB-200 dataset and comparing it to existing widely used methods, we assess how our approach handles real-world data, and showcase the merits of the semantic prototypes. We show that our method produces clear, semantically meaningful prototypes that align well with human understanding of bird species, highlighting the differentiating factors among similar species, whereas other methods fail to detect them. Through the inspection of prototypes of related species of birds that have great visual similarities, we are able to see that our method, accurately pinpoints the important features that differentiate the classes, while other methods do not. For example, in Figure~\ref{fig:Gulls} we can see two species of gulls, a ring-billed gull, and a glaucus-winged gull. The CCDs provided by our method indicate the characteristic black ring of the ring-billed as well as its black tail, and yellow eyes. When these features are juxtaposed with the CCDs provided for the glaucus-winged gull that include the characteristic pink legs and black eyes, the user is able to clearly distinguish these two species, while also understanding the characteristics of each gull. However, other widely-used methods like ProtoPNet \cite{chen2019looks}, fail to highlight these distinguishing characteristics, and indicate the wings or even the background as the prototypical patch of these classes as shown in Figure~\ref{fig:GullsProtopnet}. This can potentially be misleading and lower user trust because of the method's inability to detect the differentiating factors. 

\begin{figure}[ht]
    \centering
    \begin{subfigure}[b]{0.4\linewidth}
        \includegraphics[width=\linewidth]{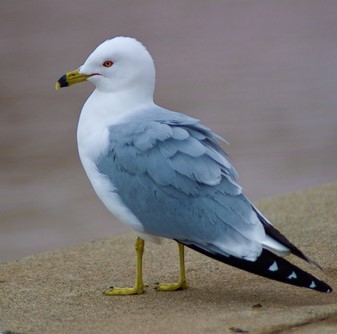}
        \caption{Ring Billed Gull}
    \end{subfigure}
    \hfill
    \begin{subfigure}[b]{0.45\linewidth}
        \includegraphics[width=\linewidth]{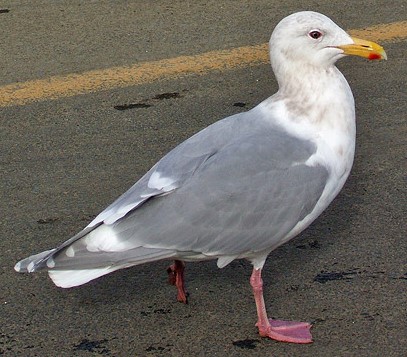}
        \caption{Glaucus Winged Gull}
    \end{subfigure}

    \caption{Two visually similar classes of gulls}
    \label{fig:Gulls}
\end{figure}

\begin{figure}[ht]
    \centering
    \begin{subfigure}[t]{0.4\linewidth}
        \includegraphics[width=\linewidth, height=2.8cm]{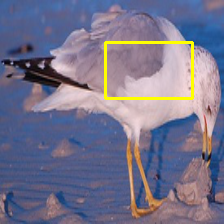} 
        \caption{Prototype for Ring Billed Gull produced by ProtoPNet \cite{chen2019looks}}
    \end{subfigure}
    \hfill
    \begin{subfigure}[t]{0.45\linewidth}
        \includegraphics[width=\linewidth, height=2.8cm]{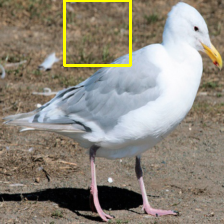} 
        \caption{Prototype for Glaucus Winged Gull produced by ProtoPNet \cite{chen2019looks}}
    \end{subfigure}

    \caption{Misleading Prototypes produced by ProtoPNet \cite{chen2019looks}}
    \label{fig:GullsProtopnet}
\end{figure}

\subsection{User Survey}
\label{sec:user}
\paragraph{\textbf{Setup}}
For the human survey,  we adopted a methodology similar to that described in \cite{vandenhende2022making, dimitriou2024structure}. The purpose of this survey is to evaluate the effectiveness of prototype methods in teaching individuals about an unfamiliar task, through two primary stages: training and testing. During the training phase, participants are exposed to prototype instances from two analogous classes within a specified method, accompanied by relevant explanations where applicable. For example, Protodash employs solely images, ProtoPNet features images with a bounding box, and our method presents images alongside textual CCD. Each participant reviews four prototypes per class, totaling eight prototypes. To mitigate bias from recognizable class names, such as ``Yellow-breasted Chat'', these names are replaced with ``Class A'' and ``Class B''. 
In the testing phase, participants are required to classify ten images from the test set as either ``Class A'' or ``Class B.'' This approach is designed to assess their ability to learn the task by simply viewing random images from the training set, without any systematic selection algorithm or additional explanations. To evaluate the generalizability of the method, the experiment employs different pairings of labels as ``Class A'' and ``Class B''. These pairs, such as Least Auklet versus Parakeet Auklet and Pelagic Cormorant versus Brandt Cormorant were selected because of their high visual similarity, which was confirmed by their high confusion rates as identified by the pretrained classifier cited in \cite{vandenhende2022making}.

Participants underwent these two phases for the following six different methodologies: only CCDs, Random Images (baseline), semantic prototypes (our method), ProtoPNet \cite{chen2019looks}, Protodash \cite{gurumoorthy2019efficient}, and ProtoPNet* (ProtoPNet prototypes along with explanations produced using the methodology introduced in \cite{nauta2021looks}). The CCDs were presented before the random images to ensure that participants had no prior knowledge about the distribution of the images, starting with only textual descriptions provided by the CCDs. These were used as a baseline for our method to evaluate the usefulness of the prototypes compared to plain semantic descriptions. Different classes of birds were randomly permutated among the different methods so that each user couldn't use prior knowledge from a previous step of the survey to classify the images of a later method. Participants were ultimately asked to indicate which prototype method they preferred and found most helpful.

The study involved 20 PhD candidates in computer science who participated voluntarily after a call for participation. Altogether, they conducted 120 tests in total. The candidates possessed no prior knowledge about bird species, which ensured an unbiased approach to the tasks presented.

\paragraph{\textbf{Results}} 
Table \ref{tab:acc-pref} presents the results, showcasing the accuracy and participant preferences for each method.
\begin{table}[b]
\caption{Accuracy in machine teaching and human preferences for each method.}
\begin{center}
\begin{tabular}{@{} l c c @{}}  
\toprule
\textbf{Method} & \textbf{Accuracy (\%)} & \textbf{Preference (\%)} \\ \midrule
Random & $77.72 \pm 8.93$ & 10 \\
ProtoPNet \cite{chen2019looks} & $77.41 \pm 9.98$ & 15 \\
ProtoPNet* \cite{nauta2021looks} & $80.03 \pm 6.83$ & 5 
\\
Protodash \cite{gurumoorthy2019efficient} & $81.37 \pm 6.49$ & 20 \\
Only CCDs &  $69.0 \pm 11.77$ & 0 \\
\textbf{ProtoSem (ours)} & $\textbf{83.12} \pm \textbf{5.19}$ & \textbf{50} \\
\bottomrule
\end{tabular}
\label{tab:acc-pref}
\end{center}
\end{table}
The results of the user survey clearly demonstrate that our method of Semantic Prototypes (ProtoSem) outperformed all other methods in terms of performance in machine teaching and user satisfaction, exhibiting the highest accuracy and the lowest standard deviation along with the highest user preference. This indicates that our approach effectively helped users focus their attention in the right direction, achieving a consistent understanding across participants.

We see a significant discrepancy between the accuracy of participants who only read the CCDs of the two classes and those who viewed actual images from the dataset. While CCDs provide essential information for differentiating each class, they alone are insufficient for users to fully grasp the necessary distinctions. Familiarity with dataset instances plays a crucial role in properly understanding how to differentiate the classes. Additionally, the high standard deviation in performance suggests that the criteria for class selection vary significantly among users. Initially, this variation might seem counter-intuitive since users are provided with the fundamental characteristics of the classes, seemingly simplifying the classification task. However, participants struggled to intuitively grasp these explanations without examples from the dataset, as interpretations of a rule such as ``The bird has a plain pattern on its head'' varied widely among users who had not seen how this characteristic manifests in actual birds.

Moreover, although adding supplementary information to each prototype intuitively appears beneficial, this is not reflected in user performance. The accuracy of users who learned with the help of explanations from ProtoPNet and ProtoPNet*, which include images along with two different types of additional information, was comparable to that of learning from randomly selected images without any explanations. Notably, ProtoPNet's performance was slightly below this baseline, with a relatively higher standard deviation, indicating that the criteria for classifying images varied considerably. ProtoPNet* showed slightly improved performance and a relatively lower standard deviation, but still very close to the baseline. 

Additionally, Protodash, which presents less information compared to other methods (except for the baseline), achieved higher performance than the preceding methods. This improvement primarily occurred because users could intuitively discern the differences between the two classes by comparing Protodash prototypes. Additionally, Protodash, which presents less information compared to other methods (except for the baseline), achieved higher performance than the preceding methods. This improvement primarily occurred because users could intuitively discern the differences between the two classes by comparing Protodash prototypes.

After each participant completed the user survey, we conducted short interviews to gather feedback and insights on the methods. Here we present some notable observations highlighted by multiple participants.
First, participants found it difficult to map the semantic information of the CCDs to the actual data when the prototypical images were not present. This underscores the importance of providing enhanced explanations in multiple formats, especially in areas where users lack expertise. Additionally, participants criticized the seemingly incorrect patches of ProtoPNet, noting that they often ignored these patches and instead identified their own patterns in the images. Many participants found the semantic explanations of ProtoPNet* unintuitive and uninterpretable because they could not relate them to the data, often choosing to disregard them.
Regarding our method, some users mentioned that the presence of the semantic description helped them identify the distinguishing characteristics of the classes, though they had to pay more attention to process all the provided information compared to methods offering only plain images. They also suggested that smaller, more focused rules would be greatly beneficial.

Regarding user preferences, it is important to note that half of the participants found our method's explanations more helpful than any of the alternatives. However, Table~\ref{tab:acc-pref} also reveals a preference for methods that offer minimal information, specifically those consisting only of images without any textual content. This is highlighted by the fact that 45\% of users identified the prototypes provided randomly, by ProtoPNet, and by Protodash as the most helpful. Interestingly, there was a stronger preference for a set of randomly selected images over the ProtoPNet* prototypes, which include images accompanied by textual explanations, even though the latter method resulted in higher user performance. Additionally, users seemed to prefer the ProtoPNet explanation, which features an image with a bounding box, despite its lower effectiveness for learning.
This highlights an important trade-off in explanation methods: informativeness versus simplicity. Some users prefer methods that contain the most useful information and can help them perform a task with careful attention, while others prefer explanations that are simple and do not require thorough investigation, even if they lead to poorer results. Therefore, it is important to keep our methods as concise as possible, avoiding unnecessary information to create explanations that are both simple and informative.

\section{Conclusions}
In this work, we have introduced a framework for producing semantic prototypes, given a dataset that is enriched with human-understandable semantic information. We have developed a methodology for computing such semantic prototypes in practice, and we have demonstrated their merits qualitatively and quantitatively. The two main takeaways from our work are that i) It is important that prototypes are accompanied by some form of explanation of ``why is this a prototype?'', which should be transparent and reliable, and ii) It is useful to compute prototypes in terms of their semantics, instead of their feature representation, and to make sure that there is as little as possible redundant information. This ensures that a user can more easily extrapolate the semantics of a class, or a cluster, compared to choosing the most representative sample in the dataset which might contain redundancies. Especially in large complex datasets, it could be challenging to find data which contains only information that causally links it to a class and as least additional information as possible, thus relying on the distribution of features could lead to less understandable prototypes, compared to relying on the semantic information. 

Regarding future extensions of our work, we have identified several key areas to be explored. Firstly, as Large Language Models (LLMs) are prevalent in both academia and industry, an interesting area to explore is the utilization of prototypes and their descriptions as a complement or enhancement to few-shot or in-context learning. Furthermore, for several natural language processing tasks, it might be useful to utilize LLMs for generating semantic descriptions, and then employing our proposed method for finding semantic prototypes in the data. A second interesting area to explore is the utilization of knowledge representation and knowledge graphs. In particular, the scale and interconnectedness of such structured data can be very useful for identifying clusters and prototypes semantically. In this regard, an extension of the algorithmic approach from \textit{sets of sets} representations to \textit{labeled directed graph} representations will provide much more expressive descriptions, which might in turn result in more understandable and informative prototypes. 
Thirdly, there is an array of domain-specific applications for the proposed methodology. An example is the domain of music, where symbolic representations, such as musical scores and notation can serve as semantic descriptions of audio recordings.  
Furthermore, besides prototypes, there are numerous other forms of explanations that could potentially benefit from utilizing the human-understandable semantic level of abstraction and could be combined with prototypes, similar to how we combine the prototypes with CCDs which are closely related to rule-based explanations. An example would be accompanying the prototypes with their counterfactual data point, along with the associated semantic descriptions. Finally, the difficulty of objectively evaluating XAI methodologies and frameworks, and reproducing results is a known issue. In the future, we plan to extend the evaluation procedure to more participants of different backgrounds, and ideally guided by disciplines of human behavior and cognition, it would be worth exploring further what a good explanation should look like.





\begin{acks}
The research project is implemented in the framework of H.F.R.I call “Basic research Financing (Horizontal support of all Sciences)” under the National Recovery and Resilience Plan “Greece 2.0” funded by the European Union –NextGenerationEU(H.F.R.I. Project Number: 15111 -
Emotional Artificial Intelligence in Music Expression).
\end{acks}

\bibliographystyle{ACM-Reference-Format}
\balance
\bibliography{bibliography.bib}


\begin{thebibliography}{45}


\ifx \showCODEN    \undefined \def \showCODEN     #1{\unskip}     \fi
\ifx \showDOI      \undefined \def \showDOI       #1{#1}\fi
\ifx \showISBNx    \undefined \def \showISBNx     #1{\unskip}     \fi
\ifx \showISBNxiii \undefined \def \showISBNxiii  #1{\unskip}     \fi
\ifx \showISSN     \undefined \def \showISSN      #1{\unskip}     \fi
\ifx \showLCCN     \undefined \def \showLCCN      #1{\unskip}     \fi
\ifx \shownote     \undefined \def \shownote      #1{#1}          \fi
\ifx \showarticletitle \undefined \def \showarticletitle #1{#1}   \fi
\ifx \showURL      \undefined \def \showURL       {\relax}        \fi
\providecommand\bibfield[2]{#2}
\providecommand\bibinfo[2]{#2}
\providecommand\natexlab[1]{#1}
\providecommand\showeprint[2][]{arXiv:#2}

\bibitem[Aamodt and Plaza(1994)]%
        {aamodt1994case}
\bibfield{author}{\bibinfo{person}{Agnar Aamodt} {and} \bibinfo{person}{Enric Plaza}.} \bibinfo{year}{1994}\natexlab{}.
\newblock \showarticletitle{Case-based reasoning: Foundational issues, methodological variations, and system approaches}.
\newblock \bibinfo{journal}{\emph{AI communications}} \bibinfo{volume}{7}, \bibinfo{number}{1} (\bibinfo{year}{1994}), \bibinfo{pages}{39--59}.
\newblock


\bibitem[Angelov and Soares(2020)]%
        {angelov2020towards}
\bibfield{author}{\bibinfo{person}{Plamen Angelov} {and} \bibinfo{person}{Eduardo Soares}.} \bibinfo{year}{2020}\natexlab{}.
\newblock \showarticletitle{Towards deep machine reasoning: a prototype-based deep neural network with decision tree inference}. In \bibinfo{booktitle}{\emph{2020 IEEE International Conference on Systems, Man, and Cybernetics (SMC)}}. IEEE, \bibinfo{pages}{2092--2099}.
\newblock


\bibitem[Arik and Pfister(2020)]%
        {arik2020protoattend}
\bibfield{author}{\bibinfo{person}{Sercan~O Arik} {and} \bibinfo{person}{Tomas Pfister}.} \bibinfo{year}{2020}\natexlab{}.
\newblock \showarticletitle{Protoattend: Attention-based prototypical learning}.
\newblock \bibinfo{journal}{\emph{Journal of Machine Learning Research}} \bibinfo{volume}{21}, \bibinfo{number}{210} (\bibinfo{year}{2020}), \bibinfo{pages}{1--35}.
\newblock


\bibitem[Augasta and Kathirvalavakumar(2012)]%
        {augasta2012reverse}
\bibfield{author}{\bibinfo{person}{M~Gethsiyal Augasta} {and} \bibinfo{person}{Thangairulappan Kathirvalavakumar}.} \bibinfo{year}{2012}\natexlab{}.
\newblock \showarticletitle{Reverse engineering the neural networks for rule extraction in classification problems}.
\newblock \bibinfo{journal}{\emph{Neural processing letters}}  \bibinfo{volume}{35} (\bibinfo{year}{2012}), \bibinfo{pages}{131--150}.
\newblock


\bibitem[Chen et~al\mbox{.}(2019)]%
        {chen2019looks}
\bibfield{author}{\bibinfo{person}{Chaofan Chen}, \bibinfo{person}{Oscar Li}, \bibinfo{person}{Daniel Tao}, \bibinfo{person}{Alina Barnett}, \bibinfo{person}{Cynthia Rudin}, {and} \bibinfo{person}{Jonathan~K Su}.} \bibinfo{year}{2019}\natexlab{}.
\newblock \showarticletitle{This looks like that: deep learning for interpretable image recognition}.
\newblock \bibinfo{journal}{\emph{Advances in neural information processing systems}}  \bibinfo{volume}{32} (\bibinfo{year}{2019}).
\newblock


\bibitem[Cima et~al\mbox{.}(2022)]%
        {cima2022separability}
\bibfield{author}{\bibinfo{person}{Gianluca Cima}, \bibinfo{person}{Federico Croce}, {and} \bibinfo{person}{Maurizio Lenzerini}.} \bibinfo{year}{2022}\natexlab{}.
\newblock \showarticletitle{Separability and its Approximations in Ontology-based Data Management}.
\newblock \bibinfo{journal}{\emph{Semantic Web}} \bibinfo{number}{Preprint} (\bibinfo{year}{2022}), \bibinfo{pages}{1--36}.
\newblock


\bibitem[Cohen et~al\mbox{.}(1996)]%
        {cohen1996metarecognition}
\bibfield{author}{\bibinfo{person}{Marvin~S Cohen}, \bibinfo{person}{Jared~T Freeman}, {and} \bibinfo{person}{Steve Wolf}.} \bibinfo{year}{1996}\natexlab{}.
\newblock \showarticletitle{Metarecognition in time-stressed decision making: Recognizing, critiquing, and correcting}.
\newblock \bibinfo{journal}{\emph{Human factors}} \bibinfo{volume}{38}, \bibinfo{number}{2} (\bibinfo{year}{1996}), \bibinfo{pages}{206--219}.
\newblock


\bibitem[Dervakos et~al\mbox{.}(2023)]%
        {dervakos2023choose}
\bibfield{author}{\bibinfo{person}{Edmund Dervakos}, \bibinfo{person}{Konstantinos Thomas}, \bibinfo{person}{Giorgos Filandrianos}, {and} \bibinfo{person}{Giorgos Stamou}.} \bibinfo{year}{2023}\natexlab{}.
\newblock \showarticletitle{Choose your data wisely: a framework for semantic counterfactuals}. In \bibinfo{booktitle}{\emph{Proceedings of the Thirty-Second International Joint Conference on Artificial Intelligence}}. \bibinfo{pages}{382--390}.
\newblock


\bibitem[Dimitriou et~al\mbox{.}(2024)]%
        {dimitriou2024structure}
\bibfield{author}{\bibinfo{person}{Angeliki Dimitriou}, \bibinfo{person}{Maria Lymperaiou}, \bibinfo{person}{Georgios Filandrianos}, \bibinfo{person}{Konstantinos Thomas}, {and} \bibinfo{person}{Giorgos Stamou}.} \bibinfo{year}{2024}\natexlab{}.
\newblock \showarticletitle{Structure Your Data: Towards Semantic Graph Counterfactuals}. In \bibinfo{booktitle}{\emph{Proceedings of the 41st International Conference on Machine Learning}} \emph{(\bibinfo{series}{Proceedings of Machine Learning Research}, Vol.~\bibinfo{volume}{235})}. \bibinfo{publisher}{PMLR}, \bibinfo{pages}{10897--10926}.
\newblock


\bibitem[Dinur and Steurer(2014)]%
        {dinur2014analytical}
\bibfield{author}{\bibinfo{person}{Irit Dinur} {and} \bibinfo{person}{David Steurer}.} \bibinfo{year}{2014}\natexlab{}.
\newblock \showarticletitle{Analytical approach to parallel repetition}. In \bibinfo{booktitle}{\emph{Proceedings of the forty-sixth annual ACM symposium on Theory of computing}}. \bibinfo{pages}{624--633}.
\newblock


\bibitem[Feige(1998)]%
        {10.1145/285055.285059}
\bibfield{author}{\bibinfo{person}{Uriel Feige}.} \bibinfo{year}{1998}\natexlab{}.
\newblock \showarticletitle{A threshold of ln n for approximating set cover}.
\newblock \bibinfo{journal}{\emph{J. ACM}} \bibinfo{volume}{45}, \bibinfo{number}{4} (\bibinfo{date}{jul} \bibinfo{year}{1998}), \bibinfo{pages}{634–652}.
\newblock
\showISSN{0004-5411}
\urldef\tempurl%
\url{https://doi.org/10.1145/285055.285059}
\showDOI{\tempurl}


\bibitem[Fellbaum(2010)]%
        {fellbaum2010wordnet}
\bibfield{author}{\bibinfo{person}{Christiane Fellbaum}.} \bibinfo{year}{2010}\natexlab{}.
\newblock \showarticletitle{WordNet}.
\newblock In \bibinfo{booktitle}{\emph{Theory and applications of ontology: computer applications}}. \bibinfo{publisher}{Springer}, \bibinfo{pages}{231--243}.
\newblock


\bibitem[Filandrianos et~al\mbox{.}(2022)]%
        {filandrianos2022conceptual}
\bibfield{author}{\bibinfo{person}{Giorgos Filandrianos}, \bibinfo{person}{Konstantinos Thomas}, \bibinfo{person}{Edmund Dervakos}, {and} \bibinfo{person}{Giorgos Stamou}.} \bibinfo{year}{2022}\natexlab{}.
\newblock \showarticletitle{Conceptual Edits as Counterfactual Explanations.}. In \bibinfo{booktitle}{\emph{AAAI Spring Symposium: MAKE}}.
\newblock


\bibitem[Gemmeke et~al\mbox{.}(2017)]%
        {gemmeke2017audio}
\bibfield{author}{\bibinfo{person}{Jort~F Gemmeke}, \bibinfo{person}{Daniel~PW Ellis}, \bibinfo{person}{Dylan Freedman}, \bibinfo{person}{Aren Jansen}, \bibinfo{person}{Wade Lawrence}, \bibinfo{person}{R~Channing Moore}, \bibinfo{person}{Manoj Plakal}, {and} \bibinfo{person}{Marvin Ritter}.} \bibinfo{year}{2017}\natexlab{}.
\newblock \showarticletitle{Audio set: An ontology and human-labeled dataset for audio events}. In \bibinfo{booktitle}{\emph{2017 IEEE international conference on acoustics, speech and signal processing (ICASSP)}}. IEEE, \bibinfo{pages}{776--780}.
\newblock


\bibitem[Gurumoorthy et~al\mbox{.}(2019)]%
        {gurumoorthy2019efficient}
\bibfield{author}{\bibinfo{person}{Karthik~S Gurumoorthy}, \bibinfo{person}{Amit Dhurandhar}, \bibinfo{person}{Guillermo Cecchi}, {and} \bibinfo{person}{Charu Aggarwal}.} \bibinfo{year}{2019}\natexlab{}.
\newblock \showarticletitle{Efficient data representation by selecting prototypes with importance weights}. In \bibinfo{booktitle}{\emph{2019 IEEE International Conference on Data Mining (ICDM)}}. IEEE, \bibinfo{pages}{260--269}.
\newblock


\bibitem[Gutiérrez-Basulto et~al\mbox{.}(2018)]%
        {ijcai2018p255}
\bibfield{author}{\bibinfo{person}{Víctor Gutiérrez-Basulto}, \bibinfo{person}{Jean~Christoph Jung}, {and} \bibinfo{person}{Leif Sabellek}.} \bibinfo{year}{2018}\natexlab{}.
\newblock \showarticletitle{Reverse Engineering Queries in Ontology-Enriched Systems: The Case of Expressive Horn Description Logic Ontologies}. In \bibinfo{booktitle}{\emph{Proceedings of the Twenty-Seventh International Joint Conference on Artificial Intelligence, {IJCAI-18}}}. \bibinfo{publisher}{International Joint Conferences on Artificial Intelligence Organization}, \bibinfo{pages}{1847--1853}.
\newblock
\urldef\tempurl%
\url{https://doi.org/10.24963/ijcai.2018/255}
\showDOI{\tempurl}


\bibitem[Hoffmann et~al\mbox{.}(2021)]%
        {hoffmann2021looks}
\bibfield{author}{\bibinfo{person}{Adrian Hoffmann}, \bibinfo{person}{Claudio Fanconi}, \bibinfo{person}{Rahul Rade}, {and} \bibinfo{person}{Jonas Kohler}.} \bibinfo{year}{2021}\natexlab{}.
\newblock \showarticletitle{This looks like that... does it? shortcomings of latent space prototype interpretability in deep networks}.
\newblock \bibinfo{journal}{\emph{arXiv preprint arXiv:2105.02968}} (\bibinfo{year}{2021}).
\newblock


\bibitem[Kim et~al\mbox{.}(2016)]%
        {kim2016examples}
\bibfield{author}{\bibinfo{person}{Been Kim}, \bibinfo{person}{Rajiv Khanna}, {and} \bibinfo{person}{Oluwasanmi~O Koyejo}.} \bibinfo{year}{2016}\natexlab{}.
\newblock \showarticletitle{Examples are not enough, learn to criticize! criticism for interpretability}.
\newblock \bibinfo{journal}{\emph{Advances in neural information processing systems}}  \bibinfo{volume}{29} (\bibinfo{year}{2016}).
\newblock


\bibitem[Kim et~al\mbox{.}(2023)]%
        {kim2023help}
\bibfield{author}{\bibinfo{person}{Sunnie~SY Kim}, \bibinfo{person}{Elizabeth~Anne Watkins}, \bibinfo{person}{Olga Russakovsky}, \bibinfo{person}{Ruth Fong}, {and} \bibinfo{person}{Andr{\'e}s Monroy-Hern{\'a}ndez}.} \bibinfo{year}{2023}\natexlab{}.
\newblock \showarticletitle{" Help Me Help the AI": Understanding How Explainability Can Support Human-AI Interaction}. In \bibinfo{booktitle}{\emph{Proceedings of the 2023 CHI Conference on Human Factors in Computing Systems}}. \bibinfo{pages}{1--17}.
\newblock


\bibitem[Krishna et~al\mbox{.}(2017)]%
        {krishna2017visual}
\bibfield{author}{\bibinfo{person}{Ranjay Krishna}, \bibinfo{person}{Yuke Zhu}, \bibinfo{person}{Oliver Groth}, \bibinfo{person}{Justin Johnson}, \bibinfo{person}{Kenji Hata}, \bibinfo{person}{Joshua Kravitz}, \bibinfo{person}{Stephanie Chen}, \bibinfo{person}{Yannis Kalantidis}, \bibinfo{person}{Li-Jia Li}, \bibinfo{person}{David~A Shamma}, {et~al\mbox{.}}} \bibinfo{year}{2017}\natexlab{}.
\newblock \showarticletitle{Visual genome: Connecting language and vision using crowdsourced dense image annotations}.
\newblock \bibinfo{journal}{\emph{International journal of computer vision}}  \bibinfo{volume}{123} (\bibinfo{year}{2017}), \bibinfo{pages}{32--73}.
\newblock


\bibitem[Li et~al\mbox{.}(2018)]%
        {li2018deep}
\bibfield{author}{\bibinfo{person}{Oscar Li}, \bibinfo{person}{Hao Liu}, \bibinfo{person}{Chaofan Chen}, {and} \bibinfo{person}{Cynthia Rudin}.} \bibinfo{year}{2018}\natexlab{}.
\newblock \showarticletitle{Deep learning for case-based reasoning through prototypes: A neural network that explains its predictions}. In \bibinfo{booktitle}{\emph{Proceedings of the AAAI Conference on Artificial Intelligence}}, Vol.~\bibinfo{volume}{32}.
\newblock


\bibitem[Liartis et~al\mbox{.}(2021)]%
        {liartis2021semantic}
\bibfield{author}{\bibinfo{person}{Jason Liartis}, \bibinfo{person}{Edmund Dervakos}, \bibinfo{person}{Orfeas Menis-Mastromichalakis}, \bibinfo{person}{Alexandros Chortaras}, {and} \bibinfo{person}{Giorgos Stamou}.} \bibinfo{year}{2021}\natexlab{}.
\newblock \showarticletitle{Semantic Queries Explaining Opaque Machine Learning Classifiers.}. In \bibinfo{booktitle}{\emph{DAO-XAI}}.
\newblock


\bibitem[Liartis et~al\mbox{.}(2023)]%
        {liartis2023searching}
\bibfield{author}{\bibinfo{person}{Jason Liartis}, \bibinfo{person}{Edmund Dervakos}, \bibinfo{person}{Orfeas Menis-Mastromichalakis}, \bibinfo{person}{Alexandros Chortaras}, {and} \bibinfo{person}{Giorgos Stamou}.} \bibinfo{year}{2023}\natexlab{}.
\newblock \showarticletitle{Searching for explanations of black-box classifiers in the space of semantic queries}.
\newblock \bibinfo{journal}{\emph{Semantic Web}} \bibinfo{number}{Preprint} (\bibinfo{year}{2023}), \bibinfo{pages}{1--42}.
\newblock


\bibitem[Mastromichalakis et~al\mbox{.}(2024)]%
        {mastromichalakis2024rule}
\bibfield{author}{\bibinfo{person}{Orfeas~Menis Mastromichalakis}, \bibinfo{person}{Edmund Dervakos}, \bibinfo{person}{Alexandros Chortaras}, {and} \bibinfo{person}{Giorgos Stamou}.} \bibinfo{year}{2024}\natexlab{}.
\newblock \showarticletitle{Rule-Based Explanations of Machine Learning Classifiers Using Knowledge Graphs}. In \bibinfo{booktitle}{\emph{AAAI Spring Symposium: MAKE}}.
\newblock


\bibitem[Miller(2019)]%
        {miller2019explanation}
\bibfield{author}{\bibinfo{person}{Tim Miller}.} \bibinfo{year}{2019}\natexlab{}.
\newblock \showarticletitle{Explanation in artificial intelligence: Insights from the social sciences}.
\newblock \bibinfo{journal}{\emph{Artificial intelligence}}  \bibinfo{volume}{267} (\bibinfo{year}{2019}), \bibinfo{pages}{1--38}.
\newblock


\bibitem[Mittelstadt et~al\mbox{.}(2019)]%
        {mittelstadt2019explaining}
\bibfield{author}{\bibinfo{person}{Brent Mittelstadt}, \bibinfo{person}{Chris Russell}, {and} \bibinfo{person}{Sandra Wachter}.} \bibinfo{year}{2019}\natexlab{}.
\newblock \showarticletitle{Explaining explanations in AI}. In \bibinfo{booktitle}{\emph{Proceedings of the conference on fairness, accountability, and transparency}}. \bibinfo{pages}{279--288}.
\newblock


\bibitem[Nauta et~al\mbox{.}(2021a)]%
        {nauta2021looks}
\bibfield{author}{\bibinfo{person}{Meike Nauta}, \bibinfo{person}{Annemarie Jutte}, \bibinfo{person}{Jesper Provoost}, {and} \bibinfo{person}{Christin Seifert}.} \bibinfo{year}{2021}\natexlab{a}.
\newblock \showarticletitle{This looks like that, because... explaining prototypes for interpretable image recognition}. In \bibinfo{booktitle}{\emph{Joint European Conference on Machine Learning and Knowledge Discovery in Databases}}. Springer, \bibinfo{pages}{441--456}.
\newblock


\bibitem[Nauta et~al\mbox{.}(2023)]%
        {nauta2023pip}
\bibfield{author}{\bibinfo{person}{Meike Nauta}, \bibinfo{person}{J{\"o}rg Schl{\"o}tterer}, \bibinfo{person}{Maurice van Keulen}, {and} \bibinfo{person}{Christin Seifert}.} \bibinfo{year}{2023}\natexlab{}.
\newblock \showarticletitle{Pip-net: Patch-based intuitive prototypes for interpretable image classification}. In \bibinfo{booktitle}{\emph{Proceedings of the IEEE/CVF Conference on Computer Vision and Pattern Recognition}}. \bibinfo{pages}{2744--2753}.
\newblock


\bibitem[Nauta et~al\mbox{.}(2021b)]%
        {nauta2021neural}
\bibfield{author}{\bibinfo{person}{Meike Nauta}, \bibinfo{person}{Ron Van~Bree}, {and} \bibinfo{person}{Christin Seifert}.} \bibinfo{year}{2021}\natexlab{b}.
\newblock \showarticletitle{Neural prototype trees for interpretable fine-grained image recognition}. In \bibinfo{booktitle}{\emph{Proceedings of the IEEE/CVF conference on computer vision and pattern recognition}}. \bibinfo{pages}{14933--14943}.
\newblock


\bibitem[Newell et~al\mbox{.}(1972)]%
        {newell1972human}
\bibfield{author}{\bibinfo{person}{Allen Newell}, \bibinfo{person}{Herbert~Alexander Simon}, {et~al\mbox{.}}} \bibinfo{year}{1972}\natexlab{}.
\newblock \bibinfo{booktitle}{\emph{Human problem solving}}. Vol.~\bibinfo{volume}{104}.
\newblock \bibinfo{publisher}{Prentice-hall Englewood Cliffs, NJ}.
\newblock


\bibitem[Ortiz(2019)]%
        {GCAI2019:Ontology_Mediated_Queries_from_Examples}
\bibfield{author}{\bibinfo{person}{Magdalena Ortiz}.} \bibinfo{year}{2019}\natexlab{}.
\newblock \showarticletitle{Ontology-Mediated Queries from Examples: a Glimpse at the DL-Lite Case}. In \bibinfo{booktitle}{\emph{GCAI 2019. Proceedings of the 5th Global Conference on Artificial Intelligence}} \emph{(\bibinfo{series}{EPiC Series in Computing}, Vol.~\bibinfo{volume}{65})}, \bibfield{editor}{\bibinfo{person}{Diego Calvanese} {and} \bibinfo{person}{Luca Iocchi}} (Eds.). \bibinfo{publisher}{EasyChair}, \bibinfo{pages}{1--14}.
\newblock
\showISSN{2398-7340}
\urldef\tempurl%
\url{https://doi.org/10.29007/jhtz}
\showDOI{\tempurl}


\bibitem[Rousseeuw and Kaufman(1987)]%
        {rdusseeun1987clustering}
\bibfield{author}{\bibinfo{person}{P Rousseeuw} {and} \bibinfo{person}{P Kaufman}.} \bibinfo{year}{1987}\natexlab{}.
\newblock \showarticletitle{Clustering by means of medoids}. In \bibinfo{booktitle}{\emph{Proceedings of the statistical data analysis based on the L1 norm conference, neuchatel, switzerland}}, Vol.~\bibinfo{volume}{31}.
\newblock


\bibitem[Rudin(2019)]%
        {rudin2019stop}
\bibfield{author}{\bibinfo{person}{Cynthia Rudin}.} \bibinfo{year}{2019}\natexlab{}.
\newblock \showarticletitle{Stop explaining black box machine learning models for high stakes decisions and use interpretable models instead}.
\newblock \bibinfo{journal}{\emph{Nature machine intelligence}} \bibinfo{volume}{1}, \bibinfo{number}{5} (\bibinfo{year}{2019}), \bibinfo{pages}{206--215}.
\newblock


\bibitem[Rymarczyk et~al\mbox{.}(2022)]%
        {rymarczyk2022interpretable}
\bibfield{author}{\bibinfo{person}{Dawid Rymarczyk}, \bibinfo{person}{{\L}ukasz Struski}, \bibinfo{person}{Micha{\l} G{\'o}rszczak}, \bibinfo{person}{Koryna Lewandowska}, \bibinfo{person}{Jacek Tabor}, {and} \bibinfo{person}{Bartosz Zieli{\'n}ski}.} \bibinfo{year}{2022}\natexlab{}.
\newblock \showarticletitle{Interpretable image classification with differentiable prototypes assignment}. In \bibinfo{booktitle}{\emph{European Conference on Computer Vision}}. Springer, \bibinfo{pages}{351--368}.
\newblock


\bibitem[Rymarczyk et~al\mbox{.}(2021)]%
        {rymarczyk2021protopshare}
\bibfield{author}{\bibinfo{person}{Dawid Rymarczyk}, \bibinfo{person}{{\L}ukasz Struski}, \bibinfo{person}{Jacek Tabor}, {and} \bibinfo{person}{Bartosz Zieli{\'n}ski}.} \bibinfo{year}{2021}\natexlab{}.
\newblock \showarticletitle{Protopshare: Prototypical parts sharing for similarity discovery in interpretable image classification}. In \bibinfo{booktitle}{\emph{Proceedings of the 27th ACM SIGKDD Conference on Knowledge Discovery \& Data Mining}}. \bibinfo{pages}{1420--1430}.
\newblock


\bibitem[Stammer et~al\mbox{.}(2021)]%
        {stammer2021right}
\bibfield{author}{\bibinfo{person}{Wolfgang Stammer}, \bibinfo{person}{Patrick Schramowski}, {and} \bibinfo{person}{Kristian Kersting}.} \bibinfo{year}{2021}\natexlab{}.
\newblock \showarticletitle{Right for the right concept: Revising neuro-symbolic concepts by interacting with their explanations}. In \bibinfo{booktitle}{\emph{Proceedings of the IEEE/CVF conference on computer vision and pattern recognition}}. \bibinfo{pages}{3619--3629}.
\newblock


\bibitem[Vandenhende et~al\mbox{.}(2022)]%
        {vandenhende2022making}
\bibfield{author}{\bibinfo{person}{Simon Vandenhende}, \bibinfo{person}{Dhruv Mahajan}, \bibinfo{person}{Filip Radenovic}, {and} \bibinfo{person}{Deepti Ghadiyaram}.} \bibinfo{year}{2022}\natexlab{}.
\newblock \showarticletitle{Making heads or tails: Towards semantically consistent visual counterfactuals}. In \bibinfo{booktitle}{\emph{European Conference on Computer Vision}}. Springer, \bibinfo{pages}{261--279}.
\newblock


\bibitem[Vazirani(2001)]%
        {vazirani2001approximation}
\bibfield{author}{\bibinfo{person}{Vijay~V Vazirani}.} \bibinfo{year}{2001}\natexlab{}.
\newblock \bibinfo{booktitle}{\emph{Approximation algorithms}}. Vol.~\bibinfo{volume}{1}.
\newblock \bibinfo{publisher}{Springer}.
\newblock


\bibitem[Wah et~al\mbox{.}(2011)]%
        {wah2011caltech}
\bibfield{author}{\bibinfo{person}{Catherine Wah}, \bibinfo{person}{Steve Branson}, \bibinfo{person}{Peter Welinder}, \bibinfo{person}{Pietro Perona}, {and} \bibinfo{person}{Serge Belongie}.} \bibinfo{year}{2011}\natexlab{}.
\newblock \showarticletitle{The caltech-ucsd birds-200-2011 dataset}.
\newblock  (\bibinfo{year}{2011}).
\newblock


\bibitem[Wan et~al\mbox{.}(2024)]%
        {wan2024interpretable}
\bibfield{author}{\bibinfo{person}{Qiyang Wan}, \bibinfo{person}{Ruiping Wang}, {and} \bibinfo{person}{Xilin Chen}.} \bibinfo{year}{2024}\natexlab{}.
\newblock \showarticletitle{Interpretable Object Recognition by Semantic Prototype Analysis}. In \bibinfo{booktitle}{\emph{Proceedings of the IEEE/CVF Winter Conference on Applications of Computer Vision}}. \bibinfo{pages}{800--809}.
\newblock


\bibitem[Wang et~al\mbox{.}(2023)]%
        {wang2023learning}
\bibfield{author}{\bibinfo{person}{Chong Wang}, \bibinfo{person}{Yuyuan Liu}, \bibinfo{person}{Yuanhong Chen}, \bibinfo{person}{Fengbei Liu}, \bibinfo{person}{Yu Tian}, \bibinfo{person}{Davis McCarthy}, \bibinfo{person}{Helen Frazer}, {and} \bibinfo{person}{Gustavo Carneiro}.} \bibinfo{year}{2023}\natexlab{}.
\newblock \showarticletitle{Learning support and trivial prototypes for interpretable image classification}. In \bibinfo{booktitle}{\emph{Proceedings of the IEEE/CVF International Conference on Computer Vision}}. \bibinfo{pages}{2062--2072}.
\newblock


\bibitem[Wang et~al\mbox{.}(2021)]%
        {wang2021interpretable}
\bibfield{author}{\bibinfo{person}{Jiaqi Wang}, \bibinfo{person}{Huafeng Liu}, \bibinfo{person}{Xinyue Wang}, {and} \bibinfo{person}{Liping Jing}.} \bibinfo{year}{2021}\natexlab{}.
\newblock \showarticletitle{Interpretable image recognition by constructing transparent embedding space}. In \bibinfo{booktitle}{\emph{Proceedings of the IEEE/CVF international conference on computer vision}}. \bibinfo{pages}{895--904}.
\newblock


\bibitem[Weinstein et~al\mbox{.}(2013)]%
        {weinstein2013cancer}
\bibfield{author}{\bibinfo{person}{John~N Weinstein}, \bibinfo{person}{Eric~A Collisson}, \bibinfo{person}{Gordon~B Mills}, \bibinfo{person}{Kenna~R Shaw}, \bibinfo{person}{Brad~A Ozenberger}, \bibinfo{person}{Kyle Ellrott}, \bibinfo{person}{Ilya Shmulevich}, \bibinfo{person}{Chris Sander}, {and} \bibinfo{person}{Joshua~M Stuart}.} \bibinfo{year}{2013}\natexlab{}.
\newblock \showarticletitle{The cancer genome atlas pan-cancer analysis project}.
\newblock \bibinfo{journal}{\emph{Nature genetics}} \bibinfo{volume}{45}, \bibinfo{number}{10} (\bibinfo{year}{2013}), \bibinfo{pages}{1113--1120}.
\newblock


\bibitem[Xue et~al\mbox{.}(2022)]%
        {xue2022protopformer}
\bibfield{author}{\bibinfo{person}{Mengqi Xue}, \bibinfo{person}{Qihan Huang}, \bibinfo{person}{Haofei Zhang}, \bibinfo{person}{Lechao Cheng}, \bibinfo{person}{Jie Song}, \bibinfo{person}{Minghui Wu}, {and} \bibinfo{person}{Mingli Song}.} \bibinfo{year}{2022}\natexlab{}.
\newblock \showarticletitle{Protopformer: Concentrating on prototypical parts in vision transformers for interpretable image recognition}.
\newblock \bibinfo{journal}{\emph{arXiv preprint arXiv:2208.10431}} (\bibinfo{year}{2022}).
\newblock


\bibitem[Zhou et~al\mbox{.}(2003)]%
        {zhou2003extracting}
\bibfield{author}{\bibinfo{person}{Zhi-Hua Zhou}, \bibinfo{person}{Yuan Jiang}, {and} \bibinfo{person}{Shi-Fu Chen}.} \bibinfo{year}{2003}\natexlab{}.
\newblock \showarticletitle{Extracting symbolic rules from trained neural network ensembles}.
\newblock \bibinfo{journal}{\emph{Ai Communications}} \bibinfo{volume}{16}, \bibinfo{number}{1} (\bibinfo{year}{2003}), \bibinfo{pages}{3--15}.
\newblock


\end{thebibliography}



\end{document}